\documentclass{article}
\usepackage{spconf,amsmath,graphicx}

\usepackage{dblfloatfix}
\usepackage{enumitem}
\usepackage[hang]{footmisc}
\setlist{nosep, leftmargin=14pt}

\usepackage{mwe} 

\title{Automated Nucleus Segmentation and Analysis in Breast Cancer}
\name{Adrian PFLEIDERER$^{a,b}$, Dominik MÜLLER$^{a,b,1}$ and Frank KRAMER$^{a}$\thanks{\textsuperscript{1} Corresponding Author: Dominik Müller\newline\hspace*{2.0mm}IT Infrastructure for Translational Medical Research\newline\hspace*{2.0mm}Alter Postweg 101, 86159 Augsburg, Germany\newline\hspace*{2.0mm}E-mail: dominik.mueller@informatik.uni-augsburg.de}}
\address{$^{a}$IT-Infrastructure for Translational Medical Research, University of Augsburg\\$^{b}$Medical Data Integration Center, Institute for Digital Medicine, University Hospital Augsburg}
\begin{document}
%
\maketitle
\begin{abstract}
\vspace{+1.5mm}
The NuCLS dataset contains over 220,000 annotations of cell nuclei in breast cancers. We show how to use these data to create a multi-rater model with the MIScnn Framework to automate the analysis of cell nuclei. For the model creation, we use the widespread U-Net approach embedded in a pipeline. This pipeline provides besides the high performance convolution neural network, several preprocessor techniques and a extended data exploration. The final model is tested in the evaluation phase using a wide variety of metrics with a subsequent visualization. Finally, the results are compared and interpreted with the results of the NuCLS study. As an outlook, indications are given which are important for the future development of models in the context of cell nuclei.
\end{abstract}
\section{Introduction}
\label{sec:intro}

Analyzing nuclei from image slides is a difficult and repetitive task in biomedical image analysis. Therefore, more and more automated processes such as deep learning through convolutional neural networks are coming to the fore. One of the main fields of application is pathology, which focuses on the detection of tumors and other abnormalities in the context of cell nuclei. A general problem with this type of automated analysis is the limited number of sufficiently well annotated training data by experts. The large-scale NuCLS study by Amgad et al. presents an efficient method that can be used to create and verify a large amount of training data from different image scans of cell nuclei \cite{NuCLS}. The next big step is to show that these annotated data could be used with a standardized pipeline framework to segment and classify cell nuclei from breast tissue. In this work, we want to show that repetitive and complex tasks in a special pathology field could be efficiently and quickly automated with a deep neural network approach.
\section{Methods}
\label{sec:format}
\begin{figure*}[t]
	\begin{minipage}[b]{1.0\linewidth}
		\centerline{\includegraphics[width=18.0cm, height=5.0cm]{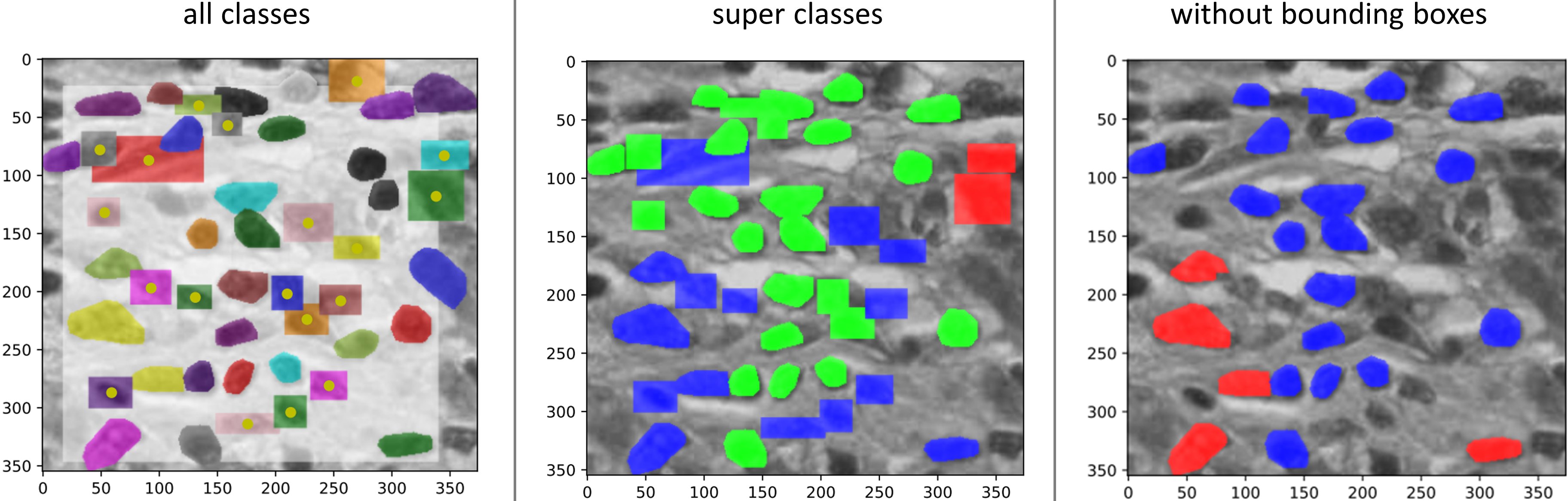}}
		\vspace{0.3cm}
		\textbf{Fig. 2.1:} Different mask representations. All instances inclusive bounding boxes (l). Tumor, stromal and sTILs as superclasses (m). Superclass representation without bounding boxes (r).\medspace\vspace{-0.2cm}
	\end{minipage}
\end{figure*}
\vspace{0.18cm}
\subsection{Data Source and Exploration}
\label{ssec:subhead}
\textbf{NuCLS Datasets.} The raw images are provided from the Cancer Genome Atlas (TCGA) program, which is supervised by the National Cancer Institute \cite{TCGA}. 
Every slide is associated with a breast cancer patient and a scan of the breast tissue. The NuCLS dataset study \cite{NuCLS} has split down the scans into 1 $mm^2$ picture tiles, which capture a region of interest. They also provide for every tile a separated mask, which annotated 13 different nucleus classes. These different region classes could merge into three superclasses like tumor, stromal and sTILs because of performance and class imbalance reasons. The first channel of the mask encodes pixel-wise the label of the specific class on the tile. The instance label of each unique nucleus segmentation could be generated with the matrix product of the second and third channel. For advanced identifications the NuCLS dataset provided for every mask a separate file, in which every class instance could be identified throw coordinates. The masks were created by 25 non-pathologists, which are guided by supervisors. They create bounding boxes or a complete segmentation for over 220,000 nuclei. This process results in two different quality datasets. The single-rater dataset contains a large number of samples (1,744), which are partially corrected and approved by the study coordinators. The multi-rater dataset contains only 53 Samples, which are annotated separately by seven experienced pathologists. The different masks for one sample are merged, which results in a high quality annotation.\newline\newline
\textbf{Class Distribution.} The tumor, stromal, and sTILs class distribution over all samples is nearly even. All samples contain the FOV-class (field of view), which consists of any excluded classes that have not been reflected into a superclass, or all areas that have not been annotated. The "other"-superclass is the only outlier with a very small amount of samples. To strengthen the superclasses the "other"-class was excluded from evaluation. The NuCLS Study also excluded this class in its model training so the scores can be easily compared in the result section.\newpage
%
\noindent\textbf{Exploration.} Before every training phase, the data were analyzed with an exploration algorithm. These calculated the mean dimensions and the class distribution over all tiles. This information is necessary for the padding routine and class imbalanced correction later in the network configuration phase. 
At the end of the exploration, different partition indices are calculated for the training, validation and evaluation phases.
\subsection{Network and MIScnn Configrations}
\label{ssec:subhead}

To create the complete network pipeline MIScnn was used \cite{miscnn21}. MIScnn is an all-in-one python library to setup a neural network pipeline system which in turn is based on the popular TensorFlow framework \cite{tensorflow}.

\subsubsection{Preprocessor}
\label{sssec:subsubhead}

For the preprocessor phase, a customized Data-IO for the MIScnn pipeline was created. The Data-IO handles the correct data input from our images and masks. Some masks and tiles have different dimensions, therefore, the oversized image is cut down and matched to the mask. A routine has also been created to optional identify and remove bounding boxes with the corresponding localization file. Finally, the mask input and their raw classes were transformed into superclasses that delivered more examples for every segmentation class.\newline\newline
\textbf{Network Input.} Three different input datasets were created for the network. The first contains only the single-rater dataset for the training, validation and evaluation phase. The second set contains also the single-rater dataset but all bounding boxes are eliminated. Both are divided into a 60-20-20 ratio for the different phases. The last Input is a combination of the original single-rater dataset and the multi-rater dataset. Whereby the complete multi-rater dataset was used for the evaluation phase. The percentage split procedure without a fold was chosen due to hardware limitations. The datasets are large enough, which is why losses in performance should be very small compared to the k-fold cross-validation procedure.\newline
\textbf{Data Augmentation.}
In addition to manual data manipulation, the MIScnn data augmentation functions were used to continue to enlarge the dataset. Besides color changes throw contrast, brightness, gamma and gaussian noise different transformations like scaling, elastic deform mirroring and rotation were applied. Each input into the neural network has been normalized with the z-score method, which calculates, for every pixel, the distance between the pixel value and the population mean with the unit of the standard deviation.\newline\newline
\textbf{Resize \& Padding Size.} Another important aspect of our preprocessing was the utilization of padding. The data tiles have different shapes, which is discussed in chapter 2.1. Tiles and masks adapted to the filter kernel are required for network input. During the data exploration, the maximum width and height over all tiles were calculated. These were used to determine the padding parameter, which had to be divisible by 2 and allowing a batch size of at least 4. This was also the resize shape for images that are larger than the average. In the context of the single-rater dataset, a maximum resolution of 796 x 830 pixels was detected, resulting in a padding and resize to 768 x 768 pixels.
\subsubsection{Network Decisions}
\label{sssec:subsubhead}

\textbf{Architecture.} The network is based on the effective standard U-Net adaptation from MIScnn. This standard Architecture was already successfully tested in many publications around the MIScnn framework and is widely used in the image segmentation domain. \cite{MIScnnExample} \cite{U-Net}\newline\newline
\textbf{Loss-Function.} A tversky variant of the reference function from Salehi et al. (2017) was used, in which the authors were able to achieve strong training improvements with 3D convolutional deep networks \cite{loss}. To further strengthen the function, we also calculated the standard cross-entropy. In order to combine both functions, the sum was used, which finally forms our loss function.\newline
\noindent\textbf{Learning Rate.} A medium-high starting value of 0.001 was used as the learning rate and decreased by a factor of 0.1 when our model reaches a plateau during training. A minimal rate of  0.00001 was defined so that the learning process remains effective. For the decrease trigger, the validation loss was monitored and a threshold from 0.0001 was specified to recognize only significant changes. After ten epochs with no significant change, the learning rate is reduced during the training process.\newline\newline
\textbf{Early Stopping.} 
If the model doesn't improve on the validation the learning phase was stopped after 30 epochs. Therefore this could be used to define the exit strategy in the learning procedure, which is better than setting an estimated maximum epoch.\newline\newline
\textbf{Batch Size.} For performance reasons, it is recommended to choose a batch size that fits in the target GPU. For the setup with a TITAN RTX graphics card (24 GB) and the padding shape specified in chapter 2.2.1, a range between 4 and 10 should be used. The best size for our setup after experimentation was 8. A value, which can be divided by 2 is recommended. The number of physical processors in GPUs often has a power of 2 distribution, which enables perfect pluralism when the batch size has been chosen in this set \cite{Pluralism}.

\subsubsection{Multi-Class Evaluation}
\label{sssec:subsubhead}

\textbf{Metrics.}  The metric collection framework, MISeval, was used to evaluate the trained model \cite{MISeval}. This enables to calculate out of the box the F1-Score, Accuracy and more. The basis of the calculation is the confusion matrix from the comparison of the model prediction and the ground truth of our test sample. All results are saved in a separate file, which is used to create our boxplot representation.\newline\newline
\textbf{Visualizations.} For every metric, a boxplot was created with the seaborn framework, which visualizes the test sample values for every class separately \cite{Seaborn}. In addition, a mean value over all samples is calculated. A routine was implemented to create a segmented image, which combines the origin tile and the predicted mask for selected samples. During the learning process, a live visualization of the progress is possible through the TensorBoard extension \cite{tensorflow}. 
\newline\newline
\begin{figure*}
	\begin{minipage}[tbp]{1.0\linewidth}
		\centerline{\includegraphics[width=17.5cm, height=4.3cm]{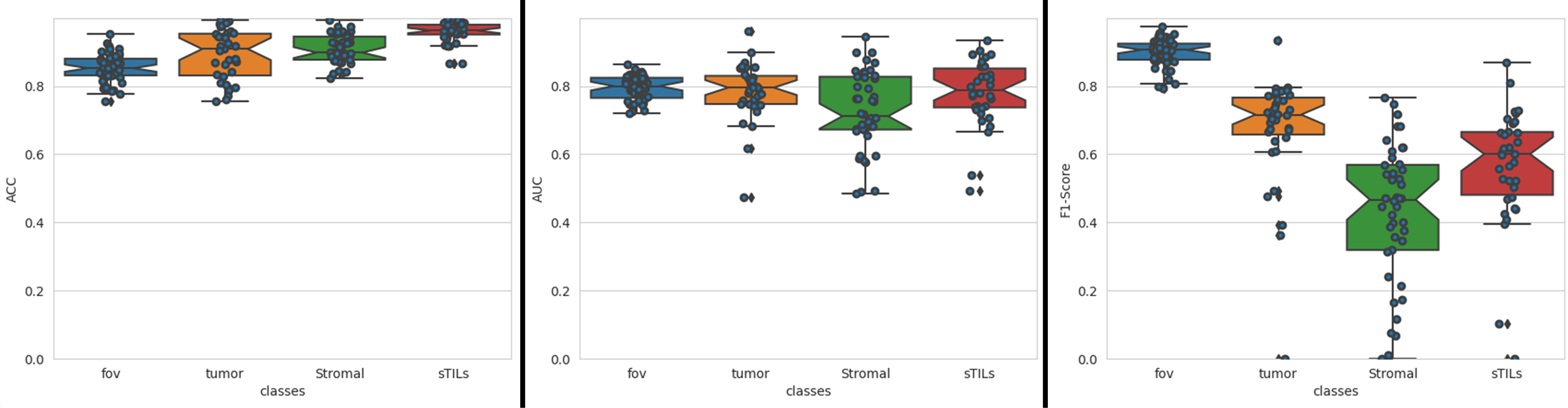}}
		\vspace{0.3cm}
		\centering
		\textbf{Fig. 3.1:} Boxplot representation of the AUC, ACC \& F1 metrics from the single \& multi-rater results.
\medspace
\vspace{-2.5mm}
	\end{minipage}
\end{figure*}

\section{Results and Discussion}
\label{sec:typestyle}

\textbf{Emerging Data Challenges.} The Data Exploration reveals that the datasets contain many inaccurate bounding boxes which have a negative impact on the evaluation task. Another problem is very rare classes and the superclass "other" (Section 2.1). To avoid extreme class imbalances and evaluation problems, these classes did not get a separate superclass and therefore they are integrated into the FOV area. The previous test attempts to include the "other"-class with the MIScnn Framework were not successful. The NuCLS study models also exclude the "other"-class from their metric analysis. Further, the samples were taken from different device types in different clinical environments. This could lead to differences in quality or unwanted influences of device characteristics on the samples. In the worst case, the model learns these peculiarities from the different recording devices, which could also have a negative impact on our scores. The NuCLS study even reinforced this problem by sorting the samples according to the devices and building per device a fold set. This approach is bypassed in this study by using a percentage split and shuffling the complete dataset before every training (Section 2.2.1).\newline\newline
\textbf{Segmentation Performance.}
Three models were created for the three different datasets discussed in Section 2.1. After around 40 - 60 epochs the training progress began to converge and no longer improved significantly. After 100 - 150 epochs the TensorFlow callback exit the training phase. The individual results of the three model evaluations are summarized and interpreted in the following sections. All scores have been rounded to one decimal place and are normalized between the range 0 to 100.
\newline\newline\textbf{Single-rater Dataset.} A large evaluation dataset with 347 samples was analyzed. This is important because the single-rater dataset has low quality annotations. A large ground truth base could compensate for possible erroneous annotations. The stromal class did not perform well in comparison to the other classes. Tumor and sTILs were able to achieve relatively better results if the outliers with a zero score were disregarded.\newline\newline
\begin{minipage}[b]{1.0\linewidth}
	\textbf{Table 3.1:} Scores of the single-rater dataset model with a 60-20-20 split and shuffled samples.\newline\newline
	\centerline{\includegraphics[width=8.5cm]{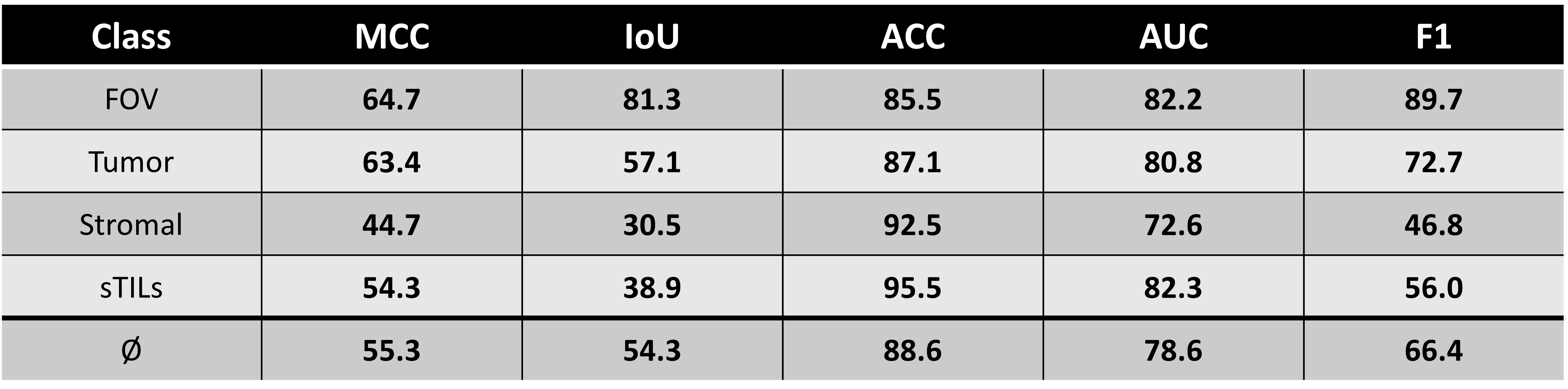}}
	\medspace
\end{minipage}
\label{key}
\textbf{Single-rater Dataset \& Multi-rater Dataset.} This model is trained and validated with 1,744 single-rater samples (low quality) and evaluated with a small set of 53 high quality annotations. The model performs in all categories similar than the exclusive single-rater dataset. This leads to the assumption that poorly to moderately annotated samples can still lead to good results if the dataset is large enough.\newline
\begin{minipage}[b]{1.0\linewidth}
	\vspace{1mm}
	\textbf{Table 3.2:} Single \& multi-rater dataset. 1,257 + 485 samples for training \& validation, 53 for evaluation (high quality).\newline\newline
	\centerline{\includegraphics[width=8.5cm]{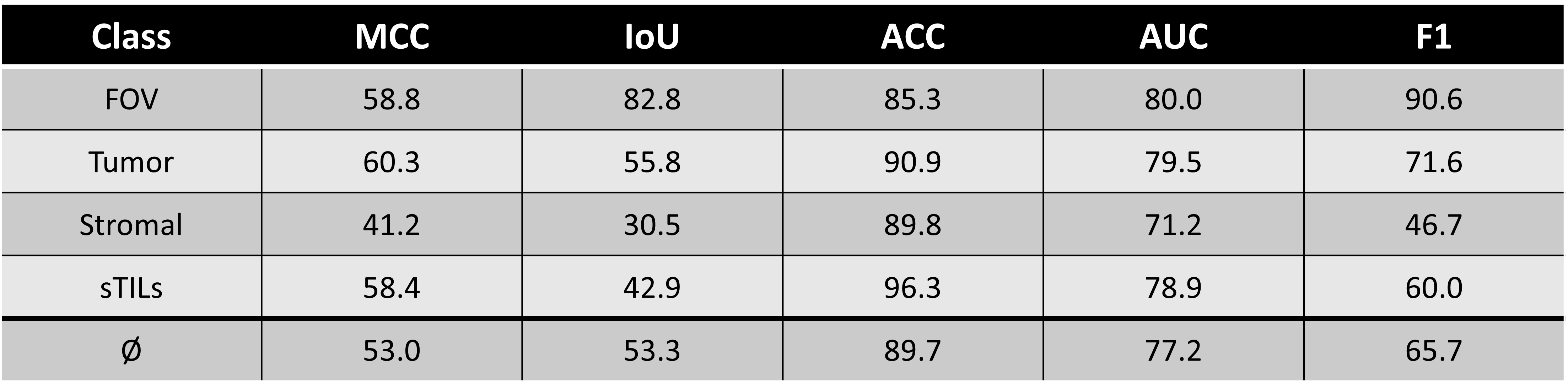}}
\end{minipage}\newline\newline
\textbf{Single-rater Dataset without Bounding Boxes.} All bounding boxes are removed without a replacement for the training. The idea was to strengthen the annotations with high quality. Unfortunately, the bounding boxes made up too much of the annotations. This resulted in a model that could not classify the data. 
Another idea could be not to remove the annotations but reduce the class weight for the bounding boxes.\newline\newline
%
\noindent\textbf{Comparison between the NuCLS and RCNN Model.} The NuCLS study provides a strongly customized model for their Dataset. They also trained a model with a general MaskRCNN \cite{MaskRCNN}. Both models are trained, validate and evaluate with the single-rater dataset. Both results should now be compared directly with our previous single-rater result. The standard MaskRCNN results are slightly worse or equal to our MIScnn Network. The NuCLS model outperforms the other standard networks. But it is important to note that the NuCLS model uses a k-fold cross-validation, which is sorted by device and recording sessions. The NuCLS network is also highly customized to the specific task.\newline\newline\vspace{+0.5cm}
\begin{minipage}[b]{1.0\linewidth}
	\textbf{Table 3.2:} Direct comparison of the different models. The overall score is calculated with the mean over all median scores per fold. The NuCLS and MaskRCNN results are extract from the work by Amgad et al. \cite{NuCLS}.
	\newline\newline
	\vspace{-1mm}\hspace{-1mm}
	\centerline{\includegraphics[width=8.6cm]{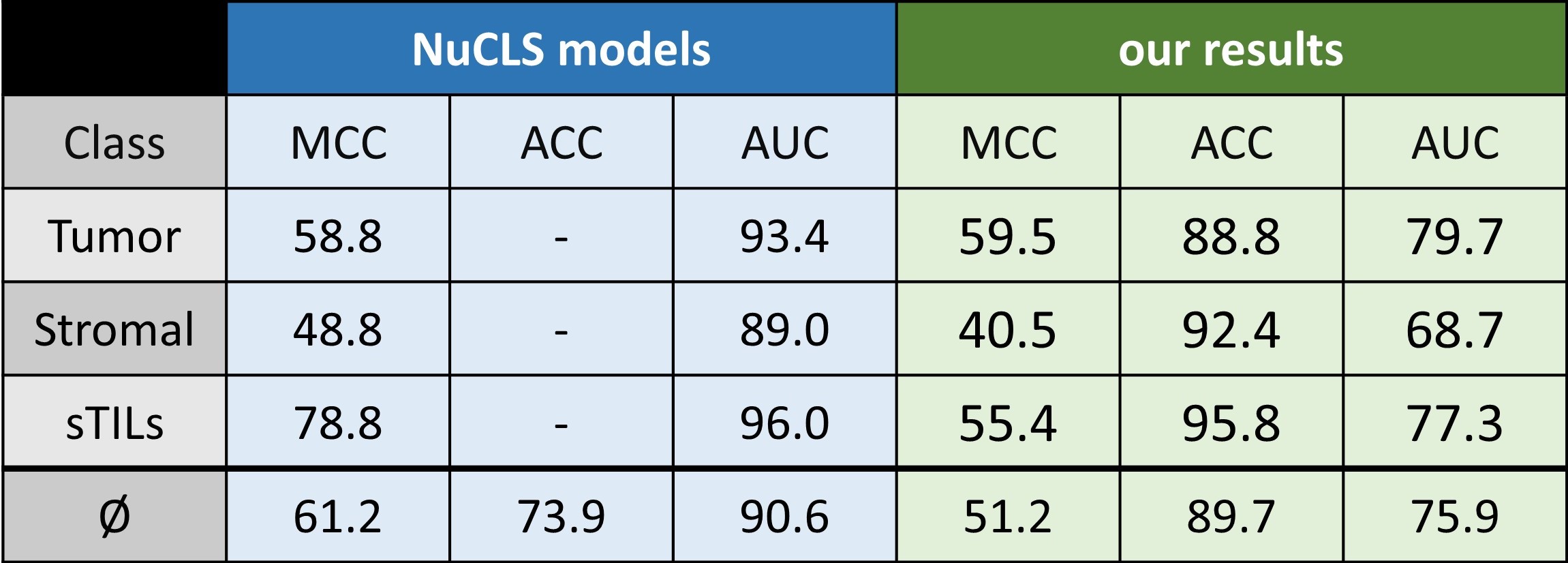}}
\end{minipage}
\noindent\textbf{Limitations.}
The single-rater model and the combination model provided solid results for the task of tumor classification of cell nuclei. Our approach to eliminating all bounding boxes did not lead to gainful results. A solution might be to not eliminate the bounding boxes completely, but only to adjust their weight per instance. In general, global weights can be introduced within the loss function to further improve the pipeline. But even with advanced adjustments to the weights, partial annotation with bounding boxes remains a limitation. A more precise annotation, especially in the multi-rater dataset, would be beneficial. The hybrid annotations types have a negative impact on the evaluation phase. To further strengthen our model, specific adjustments to our pipeline are needed. For example the usage of class probability vectors or increasing the density of region proposals, which both are present in the NuCLS study. Another important aspect is the data augmentation, which was performed in Chapter 2.2. These modifications should be used carefully and can be further adapted to the cell nucleus samples. A modification per recording device may also be useful to adapt the data sources to one another as far as possible. In the course of this, the samples could also be converted to a uniform size so that padding do not influence the result. In order to customize the model, more research is generally required in the domain of cell nuclei and their particularities.

\section{Conclusion}
\label{sec:typestyle}

Our short study showed that with a generic u-net architecture and the MIScnn pipeline, we achieved strong results in the multi-class classification of cell nuclei. The MIScnn framework makes it efficient to create initial test runs for complex models in order to approach the task. However, the created models cannot fully reach the level of an extended and adapted approach such as the NuCLS model. Nevertheless, it can be a first step in the development of these. 
MIScnn provides mature and general templates for tackling a wide range of complex medical imaging problems with an efficient collection of functions.
\newpage
\section{Appendix}
\label{sec:copyright}
The results here are in whole or part based upon data generated by the TCGA Research Network:\newline https://www.cancer.gov/tcga.\newline\newline
The code for this article was implemented in Python (platform independent) and is available under the GPL-3.0 License at the following GitHub repository:\newline https://github.com/frankkramer-lab/NuCLS.MIScnn

\section{Competing interests}
\label{sec:copyright}
This work is a part of the DIFUTURE project funded by the German Federal Ministry of Education and Research (Bundesministerium für Bildung und Forschung, BMBF) grant FKZ01ZZ1804E.

\bibliographystyle{IEEEbib}
\bibliography{strings,refs}
\end{document}